\documentclass[pdflatex,sn-mathphys-num]{sn-jnl}

\usepackage{graphicx}%
\usepackage{multirow}%
\usepackage{amsmath,amssymb,amsfonts}%
\usepackage{amsthm}%
\usepackage{mathrsfs}%
\usepackage[title]{appendix}%
\usepackage{xcolor}%
\usepackage{textcomp}%
\usepackage{manyfoot}%
\usepackage{booktabs}%
\usepackage{algorithm}%
\usepackage{algorithmicx}%
\usepackage{algpseudocode}%
\usepackage{listings}%
\usepackage{mathrsfs}

\usepackage{acronym}
\acrodef{CCE}{colon capsule endoscopy}
\acrodef{MIP}{multiple instance problem}
\acrodef{MIL}{multiple instance learning}
\acrodef{MIV}{multi-instance verification}
\acrodef{VEMA}{variance-excited multi-head attention}
\acrodef{DBA}{distance-based attention}
\acrodef{CAP}{cross-attention pooling}
\acrodef{MHSCE}{multi-head squeeze-and-excitation}
\acrodef{SimCLR}{simple framework for contrastive learning of visual representations}
\acrodef{NT-Xent}{normalized temperature-scaled cross entropy}
\acrodef{LARS}{layer-wise adaptive rate scaling}

\usepackage{booktabs}
\usepackage{makecell}

\usepackage{tikz}

\theoremstyle{thmstyleone}%
\theoremstyle{thmstyletwo}%

\theoremstyle{thmstylethree}%

\begin{document}

\title[Using Multi-Instance Learning to Identify Unique Polyps in Colon Capsule Endoscopy Images]{Using Multi-Instance Learning to Identify Unique Polyps in Colon Capsule Endoscopy Images}


\author*[1]{\fnm{Puneet} \sur{Sharma}}\email{puneet.sharma@uit.no}

\author[2]{\fnm{Kristian Dalsbø} \sur{Hindberg}}\email{kristian.d.hindberg@uit.no}

\author[3]{\fnm{Eibe} \sur{Frank}}\email{eibe.frank@waikato.ac.nz}

\author[4,5]{\fnm{Benedicte} \sur{Schelde-Olesen}}\email{benedicte.schelde-olesen@rsyd.dk}

\author[4,5]{\fnm{Ulrik} \sur{Deding}}\email{ulrik.deding@rsyd.dk}


\affil*[1]{\orgdiv{Department of Automation and Process Engineering, Machine Learning Group}, \orgname{UiT -- The Arctic University of Norway}, \orgaddress{\city{Tromsø}, \country{Norway}}}

\affil[2]{\orgdiv{Department of Physics and Technology, Machine Learning Group}, \orgname{UiT -- The Arctic University of Norway}, \orgaddress{\city{Tromsø}, \country{Norway}}}

\affil[3]{\orgdiv{Department of Computer Science}, \orgname{University of Waikato}, \orgaddress{\city{Hamilton}, \country{New Zealand}}}

\affil[4]{\orgdiv{Department of Surgery}, \orgname{Odense University Hospital}, \orgaddress{\city{Odense}, \country{Denmark}}}

\affil[5]{\orgdiv{Department of Clinical Research}, \orgname{University of Southern Denmark}, \orgaddress{\city{Odense}, \country{Denmark}}}


\abstract{Identifying unique polyps in colon capsule endoscopy (CCE) images is a critical yet challenging task for medical personnel due to the large volume of images, the cognitive load it creates for clinicians, and the ambiguity in labeling specific frames. This paper formulates this problem as a multi-instance learning (MIL) task, where a query polyp image is compared with a target bag of images to determine uniqueness. We employ a multi-instance verification (MIV) framework that incorporates attention mechanisms, such as variance-excited multi-head attention (VEMA) and distance-based attention (DBA), to enhance the model's ability to extract meaningful representations. Additionally, we investigate the impact of self-supervised learning using SimCLR to generate robust embeddings. Experimental results on a dataset of 1912 polyps from 754 patients demonstrate that attention mechanisms significantly improve performance, with DBA L1 achieving the highest test accuracy of 86.26\% and a test AUC of 0.928 using a ConvNeXt backbone with SimCLR pretraining. This study underscores the potential of MIL and self-supervised learning in advancing automated analysis of Colon Capsule Endoscopy images, with implications for broader medical imaging applications.}


\maketitle

\section{Introduction}

The \ac{MIP} occurs when the training examples are ambiguous~\cite{MIPdef}. In such cases, a single example object can be represented by multiple alternative feature vectors (instances), and it is unclear which of the feature vectors are responsible for the object's observed classification~\cite{MIPdef}.

\Ac{MIL} builds on the concept of \ac{MIP} and is a type of machine learning framework where each training example is represented as a bag containing multiple instances, rather than a single feature vector~\cite{MIL_def, ilse2018attention}. In this framework, the label is assigned to the entire bag, not to individual instances. The goal of multi-instance learning is to predict the label of unseen (both positive and negative) bags by analyzing the instances within them~\cite{MIL_def, ilse2018attention}. 

In this paper, we apply a recently introduced variant of \ac{MIL} called  \ac{MIV}~\cite{xu2025multiple}. In~\ac{MIV}, a training example consists of a bag of instances {\em plus} a single query instance. Such a training example is labeled positive if the query instance matches some instances in the target bag and negative otherwise. In the application considered in this paper, this is useful because it allows matching a query image against a bag of images when only some of the images in the bag show a clear relationship to the query--even a single clear match should suffice.  

\Ac{CCE} is a non-invasive diagnostic procedure used to examine the colon for abnormalities, such as polyps, inflammation, or signs of colorectal cancer~\cite{spada2015colon,  eliakim2009prospective, pasha2018applications}. It involves swallowing a small, pill-sized capsule (PillCam COLON 2 from Medtronic or similar from other companies) equipped with a tiny camera, light source, and transmitter~\cite{eliakim2006evaluation, eliakim2009prospective, hosoe2021current}. As the capsule travels through the digestive tract, it captures thousands of images of the colon, which are transmitted to an external recording device for subsequent analysis. \ac{CCE} achieves polyp detection rates comparable to those of optical colonoscopy while also offering enhanced cost-effectiveness~\cite{spada2011second, pasha2018applications}, and hence is a suitable candidate for colorectal screening. The cost-effectiveness of the \ac{CCE} approach stems from the overall gains achieved throughout the pipeline, particularly in scenarios where clinicians are involved and patients from remote locations are required to travel to a healthcare facility. By minimizing the need for travel, the approach significantly reduces associated expenses such as transportation, accommodation, and time lost from work for patients. This not only alleviates the financial burden on patients, but also optimizes resource allocation for healthcare providers.

In \ac{CCE}, a single video fragment corresponding to a particular polyp yields an image sequence (bag) that contains multiple frames (instances). The question we consider in this paper is whether a given other frame extracted from the original video, the query instance, can be successfully matched with the bag if it shows the same polyp as the images in the bag. Conversely, machine learning should reject a match if the query image does not pertain to the same polyp. This is a good fit for the \ac{MIV} paradigm because not all frames will necessarily show a clear view of the relevant polyp. In the training data, the label for a particular query-bag pair depends on which specific frames (instances) are clearly associated with the polyp concerned and thus can be matched with the query. \Ac{MIV} helps to handle this ambiguity by focusing on identifying the instances within the bag that match the query best. 

It is important to note that as the colon capsule can move back and forth inside the colon, this can result in hundreds of detections of polyps at the end of a \Ac{CCE} and comparing individual images for uniqueness can be a time consuming task, which can be mitigated by using \ac{MIL}. For instance, if we assume $N$ polyp detections obtained at the end of a screening, then comparing $N$ images for uniqueness requires comparing each image with every other image with the number of comparisons $N(N-1)/2$ and computational complexity of $O(N^{2})$. By employing \Ac{MIV} and dividing the images into bags, one query image can be compared against multiple ($k$) target images simultaneously. In this manner, With careful bag organization, the number of required comparisons can be reduced to approximately $N(N-1)/(2k)$, achieving a computational gain of up to $k\times$ compared to exhaustive pairwise comparison.

\Ac{MIL} is applied in a large number of medical applications~\cite{MIL_review_apps} ranging from diabetic retinopathy in retinal images~\cite{schlegl2015predicting}, cancer types in stomach computed tomography (CT) images~\cite{Li2015}, to histology images~\cite{Zhao2006}. In this context, the application of \ac{MIL} to \ac{CCE} images is a relatively new direction, and the application of the \ac{MIV} variant to medical imagery is also novel.  

The key contributions of this paper lie in addressing the challenging task of identifying unique polyps in \ac{CCE} images by formulating it as a \ac{MIL} problem. The study leverages a \ac{MIV} framework that uses advanced attention mechanisms, to enhance the model's ability to derive significant representations from ambiguous and complex image data. Furthermore, the paper explores the integration of self-supervised learning through SimCLR to generate robust embeddings, which significantly improve the performance of the \ac{MIV} framework.

The rest of the paper is organized as follows: Section~\ref{sec:data} describes the dataset used in this study, including its structure, characteristics, and the preprocessing steps undertaken to prepare it for analysis. Section~\ref{sec:method} outlines the methodology, detailing the \ac{MIV} framework, the attention mechanisms employed, and the integration of self-supervised learning using SimCLR. Section~\ref{sec:res} presents the experimental results, including performance comparisons across different models and attention mechanisms, as well as an analysis of the impact of SimCLR pretraining. In addition, we discuss the challenges encountered and explore the broader implications of the proposed approach. Finally, we conclude the paper by summarizing the key contributions.

\section{Data}
\label{sec:data}

The data used in our study is based on the CareForColon2015~\cite{kaalby2020colon, baatrup2025choice} trial conducted in the Danish colorectal cancer screening programme, which used the PillCam\texttrademark\ Colon 2 (Medtronic, USA)~\citep{Negreanu2013} camera capsule. We have access to 2,780 polyps from 853 patients from the CareForColon2015 study for which clinicians from the Department of Surgery of Odense University Hospital have exported up to five images per polyp. These five images correspond to 1) first partial, 2) first full, 3) best full, 4) last full, and 5) last partial image of the same polyp. This means that image 1 and 5 might just partially show the polyp. These are images from the same passing, so we know that there are no cases of the same polyp being seen at a significant later passage time. When we restrict the set to those polyps that the clinicians were able to export all these five images, we end up with 1912 polyps from 754 patients. These include 263 patients with one polyp, 204 patients with 2 polyps, 113 patients with 3 polyps, 77 patients with 4 polyps, and the rest with 5 or more polyps. Each set of five images from a polyp is given a unique patient ID and a unique polyp number. 
\begin{figure}[htbp]
\centering
\includegraphics[width=0.85\textwidth]{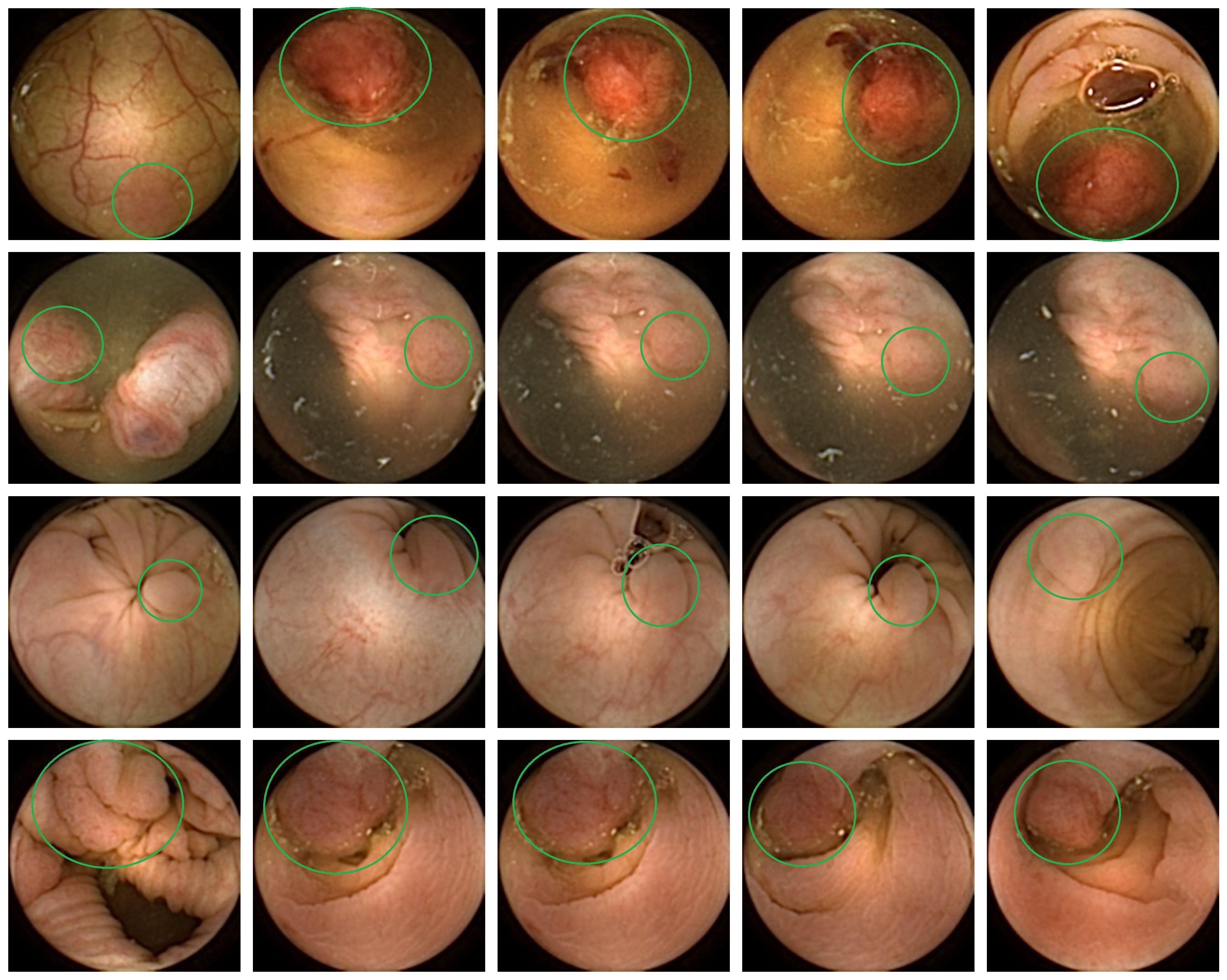}
\caption{Positive exemplar images with each row showing five images of a unique polyp, where in each row, the first image corresponds to first partial and fifth image is the last full view of the same polyp. The polyps have been marked in green for each image.}
\label{fig:example_pos}
\end{figure}

For \ac{MIV}, we construct both positive and negative exemplar pairs from this multi-patient polyp image dataset. Positive exemplars are constructed based on the five images from a single polyp, with one of the second, third, or fourth image designated as the query and the remaining four images forming the target bag. Images from five example polyps are shown in Figure~\ref{fig:example_pos}. Negative exemplars are constructed using a more sophisticated strategy: if a patient has multiple polyps, the query image is selected from one of those polyps at random, while target images are sampled from other polyps of the same patient; if a patient has only one polyp, target images are drawn from different patients entirely.

To ensure robust evaluation and prevent patient-level data leakage, we implement a patient-level stratification across all data splits. The dataset is first divided into test and training/validation sets at the patient level, with 20\% of unique patients reserved for testing. The remaining patients are then partitioned into $k$-fold cross-validation splits ($k$=10), where each fold maintains complete patient separation: no patient appears in both training and validation sets of the same fold, and test patients are excluded from all folds entirely. This stratification strategy ensures that the model's generalization is evaluated on genuinely unseen patients providing a more realistic assessment of clinical deployment performance.

\section{Method}
\label{sec:method}

In this section, we briefly discuss the details of the \ac{MIV} approach and the \ac{MIV} with SimCLR approach used for our experiments.

\subsection{Multi-instance Verification}

\usetikzlibrary{shapes.geometric} 

\begin{figure}[htbp]
\centering

\begin{tikzpicture}
    \tikzstyle{box} = [draw=black, thick, minimum width=0.75cm, minimum height=0.75cm]
    \tikzstyle{boxtext} = [font=\bfseries, anchor=south]
    \tikzstyle{arrow} = [line width=1mm, ->, >=stealth]

    \tikzstyle{encoderbox} = [draw=black, thick, trapezium, trapezium angle=75, minimum width=0.5cm, minimum height=2cm]
    \tikzstyle{encodertext} = [font=\bfseries]
    \tikzstyle{embedding} = [draw=black, thick, minimum width=0.5cm, minimum height=1cm, align=center]
    
    \node[box] (queryBox) at (0,0) {};
    \node[boxtext] at (queryBox.north) {Query};
    \node[box] (targetBox) at (0.50,-2.50) {};
    \node[box] (targetBox) at (0.25,-2.25) {};
    \node[box] (targetBox) at (0,-2.0) {};

    \node[boxtext] at (targetBox.north) {Target Bag};
    \node[encoderbox,rotate=270] (encoderBox) at (3,-1) {};
    
    \node[align=center][font=\bfseries] at (3,1.25) {Feature \\ Extraction};
    \node[align=center] at (3,-1) {Pre-trained \\ or \\ SimCLR};    
    \node[align=center][font=\bfseries] at (6,1.25) {Multi-Head \\ Attention};
     \node[align=center] at (6,-1) {VEMA \\ DBA L1 \\ DBA L2 }; 
    \node[align=center][font=\bfseries] at (9.75,1.25) {Siamese \\ Network};
    \node[align=center][font=\bfseries] at (8,-0.25) {$V_Q$};
    \node[align=center][font=\bfseries] at (8,-1.75) {$V_T$};
    \node[align=center][font=\bfseries] at (11.5,-1) {True \\ or \\ False};
    \draw [arrow](1,0) -- (2,0);
    \draw [arrow](1.25,-2.0) -- (2,-2.0);
    \draw [arrow](4,-0.5) -- (5,-0.5);
    \draw [arrow](4,-1.5) -- (5,-1.5);
    \draw [arrow](7,-0.25) -- (7.75,-0.25);
    \draw [arrow](7,-1.75) -- (7.75,-1.75);
    \draw [arrow](8.25,-0.25) -- (9,-0.25);
    \draw [arrow](8.25,-1.75) -- (9,-1.75);
    \draw [arrow](10.5,-1) -- (11.25,-1);
    \draw[thick] (7,-2.5) rectangle (5,0.5);    
    \node[embedding] (VQ) at (8.0,-0.25) {};
    \node[embedding] (VT) at (8.0,-1.75) {};
    \draw[thick] (9,-2.5) rectangle (10.5,0.5);

\end{tikzpicture}
\caption{A diagram showing the framework used for \ac{MIV} in this study. The feature extraction is performed for both the query image and the target bag. Multi-head attention mechanisms are used to extract representations from the query, denoted by $V_Q$, and the target set, denoted by $V_T$. These representations are then used by a Siamese network for the classification task.}
\label{fig:framework_for_miv}
\end{figure}
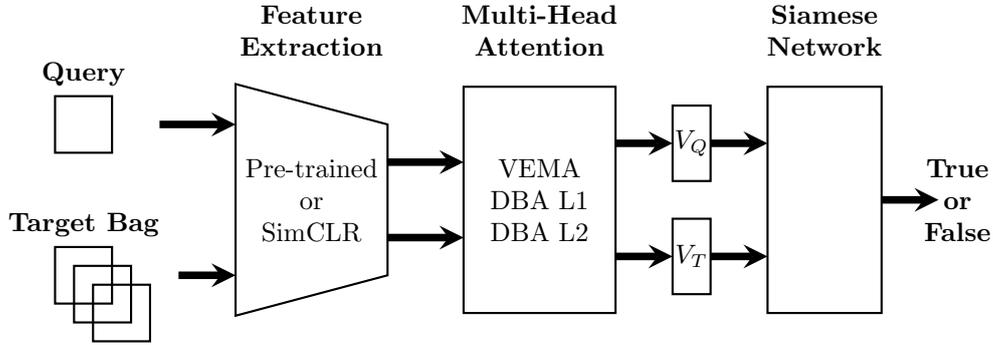

For our implementation, we use a Siamese neural network designed for \ac{MIV}, based on~\cite{xu2025multiple}. The network consists of a shared feature transformation that projects input embeddings extracted from a pretrained model (e.g., EfficientNet~\cite{efficientnet}, ResNet~\cite{resnet}, ConvNeXt~\cite{convnext}, or ViT~\cite{ViT}) through a two-layer fully connected network with batch normalization, ReLU activation, dropout regularization, and group normalization, mapping inputs to an embedding vector of 512 dimensions. These vectors are subsequently processed by multi-head \ac{CAP}, implemented using one of three possible attention functions, followed by similarity calculations~\cite{xu2025multiple}. The key mechanism here is \ac{CAP}, which employs cross-attention to use information from the query vector to adaptively pool the four vectors in the target bag into a single vector. To investigate the effect of using multiple attention heads, we also include a version of the architecture that performs single-head attention-based pooling in our experiments.  

As shown in Figure~\ref{fig:framework_for_miv}, the feature extraction is performed for both query image and a target bag (the number of images used in the target set is 4). This step is followed by using multi-head attention, which extracts representations of the query, yielding a vector $V_Q$, and the target set, yielding a vector $V_T$. These representations are then used by a Siamese network for the classification task. 

\subsubsection{Different Attention Mechanisms}


\Ac{VEMA}~\cite{xu2025multiple} computes attention scores by gating query vectors through variance-derived transformations of the target set, incorporating learnable projection matrices $R$ and $S$ to modulate query representations based on target variance. \Ac{DBA}~\cite{xu2025multiple} calculates attention weights using either L1 or L2 distances between query and target embeddings, normalized by learnable beta parameters and theoretical constants derived from the expected distance distributions. \Ac{DBA} is considered a simpler alternative to \ac{VEMA}~\cite{xu2025multiple}. \Ac{MHSCE} applies channel-wise recalibration through global average pooling and two fully connected layers, generating excitation vectors that modulate both the attended target representations and original query embeddings. Additionally, as a baseline, we use no-attention mechanisms using mean or max pooling, thus enabling a direct comparison against attention-free aggregation strategies.

\subsubsection{Training and Hyperparameters}

We employ $k$-fold cross-validation with patient-level stratification. Each fold is trained for up to 50 epochs using binary cross-entropy loss optimized via RMSprop (in line with~\cite{xu2025multiple}) with an initial learning rate of 0.0001, which is reduced using a learning rate scheduler. The training procedure incorporates early stopping (with a patience of 10 epochs) regarding the validation loss, which seeks to identify the optimal model checkpoint and prevent overfitting.


The experiments evaluate multiple architectural configurations by considering a grid over attention types (\ac{VEMA}, \ac{DBA}-L1, \ac{DBA}-L2, no-attention with mean pooling, no-attention with max pooling) and numbers of attention heads (h = 1, 2, 4, 8, 16). For each configuration, models are trained across all $k = 10$ folds, with the best-performing fold model (associated with the least loss) subsequently evaluated on the held-out test set. All predictions utilize a fixed decision threshold of 0.5 (model outputs probabilities are in the range [0,1]) for binary classification, ensuring consistent evaluation across configurations and eliminating threshold optimization as a confounding variable.

\subsection{SimCLR and MIV}
In order to investigate whether the performance of the \ac{MIV} models used in this paper can benefit from a contrastive learning approach. We also consider using \ac{SimCLR}~\cite{chen2020simple} to generate embeddings from polyp images from the PillCam data.

We implement a \ac{SimCLR} architecture consisting of three primary components: a backbone encoder, a projection head, and a contrastive loss function. The backbone encoder utilizes pre-trained deep learning models (EfficientNet~\cite{efficientnet}, ResNet~\cite{resnet}, ConvNeXt~\cite{convnext}, ViT~\cite{ViT}) to extract high-dimensional feature representations from input images. The projection head is implemented as a two-layer perceptron that transforms these backbone features through a hidden layer of 2048 dimensions before projecting to a final 512-dimensional embedding space. For details, please see Table~\ref{tab:sim_clr_details}. This projection head incorporates batch normalization and employs ReLU activation. It applies L2 normalization to the final projections, ensuring all embeddings reside on the unit hypersphere, where angular distances effectively capture semantic similarity. The contrastive learning objective is realized through the \ac{NT-Xent} loss~\cite{robinson2020contrastive}, which maximizes agreement between differently augmented views of the same image while minimizing agreement with augmented views of other images within the batch. 
\begin{table}[htbp]
\centering
\caption{Details of the models and parameters used for SimCLR experiments. All experiments run 200 epochs and a batch size of 64.}
\label{tab:sim_clr_details}
\begin{tabular}{lccc}
\toprule
\makecell{Pretrained \\ base model} & \makecell{Embedding \\ dimension} & \makecell{Projection \\ dimension} & Optimizer\\
\midrule
ViT             &  768 & 512 & AdamW \\
ResNet-50        & 2048 & 512 & LARS \\
Efficientnet-B5 & 2048 & 512 & LARS \\
ConvNeXt        & 1024 & 512 & LARS \\
\bottomrule
\end{tabular}
\end{table}

\subsubsection{Training SimCLR}

The training procedure for the feature extractor based on \ac{SimCLR} is tailored to the specific characteristics of different backbone architectures. For convolutional neural network backbones, we employ the \ac{LARS}~\cite{LARSopt} optimizer with a base learning rate of 0.3, scaled proportionally to the batch size, combined with minimal weight decay of 1e-6 and momentum of 0.9. For the Vision Transformer backbone, the implementation uses the AdamW optimizer~\cite{AdamWOpt} with a lower base learning rate of 3e-4 and stronger regularization through weight decay of 0.05. The learning rate schedule incorporates a warmup strategy~\cite{goyal2017accurate, vaswani2017attention, He_2019_CVPR} for 10 epochs, during which the learning rate linearly increases from zero to the base value, followed by cosine annealing decay~\cite{loshchilov2016sgdr, He_2019_CVPR} to a minimum learning rate of 1e-5 over the remaining training duration of up to 200 epochs. Data augmentation applies multiple stochastic transformations including random resized crops, horizontal and vertical flips, rotations, color jittering, random grayscale, and Gaussian blur. Early stopping with patience of 20 epochs is used. 


\section{Results}
\label{sec:res}
Based on the results shown in Tables~\ref{tab:resnet_pre},~\ref{tab:resnet_simclr},~\ref{tab:vit_pre},~\ref{tab:vit_simclr},~\ref{tab:efficientnet_pre},~\ref{tab:efficientnet_simclr},~\ref{tab:ConvNeXt_pre},~\ref{tab:ConvNeXt_simclr}, we can observe that the baseline models without attention mechanisms i.e., no attention(mean) and no attention(max) consistently achieve moderate performance across all architectures, with validation accuracies ranging from approximately 75.6\% to 81.7\% and test accuracies between 74.6\% and 82.9\%. Similarly, AUC scores on the validation data range from approximately 0.82\% to 0.9\% and those on the test data range from 0.82\% to 0.91\%. These baselines establish that both mean and max pooling provide reasonable starting points. Note that neither consistently outperforms the other across different backbone architectures.

Attention mechanisms substantially improved model performance across all tested architectures. \ac{VEMA}-based configurations demonstrate strong improvements, particularly when using multiple attention heads. For instance, in Tables~\ref{tab:resnet_pre}, with the pretrained ResNet-50 backbone, \ac{VEMA} with 16 heads achieves a test accuracy of 83.55\% and test AUC of 0.924, representing significant gains over the no-attention baseline. In a similar manner (in Table~\ref{tab:vit_pre}), we notice that the ViT backbone shows steady improvement from single-head to multi-head configurations, reaching 81.05\% test accuracy and test AUC of 0.901 with 16 heads. Similar trends are observed for test accuracies and test AUCs  for EfficientNet and ConvNeXt backbones, respectively, as shown in Tables~\ref{tab:efficientnet_pre}, and~\ref{tab:ConvNeXt_pre}.

The \ac{DBA} mechanisms, implemented with both L1 and L2 distance metrics, shows competitive and often superior performance compared to VEMA. Notably, DBA consistently achieves better results across different number of attention heads. For the SimCLR framework with ConvNeXt backbone, DBA L1 with 2 heads achieves the highest test accuracy of 86.26\% and a test AUC of 0.928. The choice between L1 and L2 distance metrics appears architecture-dependent, with neither consistently dominating across all experiments performed in this paper.

The integration of self-supervised learning through SimCLR pretraining generally enhances performance across architectures. SimCLR-pretrained models often achieved higher validation and test accuracies compared to their standard pretrained counterparts. For example, the EfficientNet backbone with SimCLR pretraining and DBA L1 with 2 heads reaches approx. 84.82\% validation accuracy, 83.46\% test accuracy and test AUC of 0.926 , outperforming the standard pretrained EfficientNet configuration. This suggests that the learned representations from contrastive learning provide a good foundation for attention-based classification.

Across all tested architectures, the optimal number of attention heads varies, but configurations with 2 to 8 heads frequently yielded the best results. Increasing to 16 heads sometimes improves performance but occasionally leads to slight degradation (see Tables~\ref{tab:resnet_simclr} and~\ref{tab:ConvNeXt_pre}), suggesting that excessive attention capacity may introduce noise or overfitting. The ConvNeXt architecture with SimCLR pretraining achieved the highest overall performance metrics, indicating that modern convolutional architectures combined with self-supervised learning and attention mechanisms represent a particularly effective combination for this classification task.

\begin{table}[ht]
\centering
\caption{Validation and test performance across different attention mechanisms for pretrained ResNet-50 model. Please note that validation accuracy and validation AUC values (mean $\pm$ one standard deviation) are based on $k=10$ folds.}
\label{tab:resnet_pre}
\begin{tabular}{lcccc}
\toprule
Experiment & Val Acc (\%) & Test Acc (\%) & Val AUC & Test AUC \\
\midrule
No attention(mean) & 79.19$\pm$2.65 & 77.72 & 0.871$\pm$0.023 & 0.852 \\
No attention(max) & 78.85$\pm$3.89 & 79.27 & 0.863$\pm$0.037 & 0.881 \\
VEMA(h=1) & 79.59$\pm$1.67 & 76.68 & 0.866$\pm$0.019 & 0.851 \\
VEMA(h=2) & 82.63$\pm$1.72 & 82.64 & 0.890$\pm$0.019 & 0.897 \\
VEMA(h=4) & 82.68$\pm$1.67 & 82.51 & 0.892$\pm$0.023 & 0.902 \\
VEMA(h=8) & 82.92$\pm$1.54 & 80.57 & 0.896$\pm$0.019 & 0.897 \\
VEMA(h=16) & \underline{83.36$\pm$2.38} & \underline{83.55} & 0.898$\pm$0.022 & \underline{0.924} \\
DBA L1(h=1) & 79.14$\pm$2.62 & 79.15 & 0.866$\pm$0.034 & 0.879 \\
DBA L1(h=2) & 83.01$\pm$2.96 & 80.70 & 0.896$\pm$0.028 & 0.875 \\
DBA L1(h=4) & 83.24$\pm$1.99 & 82.25 & 0.900$\pm$0.016 & 0.886 \\
DBA L1(h=8) & 82.28$\pm$2.03 & 82.38 & 0.899$\pm$0.021 & 0.907 \\
DBA L1(h=16) & 82.89$\pm$1.88 & 83.03 & 0.895$\pm$0.024 & 0.902 \\
DBA L2(h=1) & 79.67$\pm$2.68 & 78.37 & 0.868$\pm$0.021 & 0.871 \\
DBA L2(h=2) & 82.71$\pm$2.45 & 83.16 & 0.892$\pm$0.028 & 0.902 \\
DBA L2(h=4) & 82.64$\pm$2.20 & 83.16 & 0.893$\pm$0.020 & 0.894 \\
DBA L2(h=8) & 83.29$\pm$2.89 & 82.51 & \underline{0.902$\pm$0.025} & 0.902 \\
DBA L2(h=16) & 82.92$\pm$2.24 & \underline{83.55} & 0.897$\pm$0.019 & 0.900 \\
\bottomrule
\end{tabular}
\end{table}

\begin{table}[htbp]
\centering
\caption{Validation and test performance across different attention mechanisms for SimCLR with pretrained ResNet-50 backbone}
\label{tab:resnet_simclr}
\begin{tabular}{lcccc}
\toprule
Experiment & Val Acc (\%) & Test Acc (\%) & Val AUC & Test AUC \\
\midrule
No attention(mean) & 79.85$\pm$3.53 & 80.53 & 0.883$\pm$0.029 & 0.886 \\
No attention(max) & 80.47$\pm$3.26 & 78.12 & 0.884$\pm$0.031 & 0.879 \\
VEMA(h=1) & 80.37$\pm$2.91 & 81.17 & 0.884$\pm$0.023 & 0.896 \\
VEMA(h=2) & 84.09$\pm$3.18 & 83.97 & 0.901$\pm$0.030 & 0.910 \\
VEMA(h=4) & 84.36$\pm$3.02 & 83.21 & 0.899$\pm$0.026 & 0.891 \\
VEMA(h=8) & 83.71$\pm$3.20 & 83.59 & 0.899$\pm$0.024 & 0.895 \\
VEMA(h=16) & 82.76$\pm$2.69 & \underline{84.99} & 0.900$\pm$0.026 & 0.915 \\
DBA L1(h=1) & 80.09$\pm$3.10 & 79.77 & 0.882$\pm$0.027 & 0.881 \\
DBA L1(h=2) & 84.43$\pm$3.05 & 83.72 & 0.905$\pm$0.025 & 0.890 \\
DBA L1(h=4) & 83.35$\pm$2.46 & 84.61 & 0.898$\pm$0.025 & \underline{0.925} \\
DBA L1(h=8) & 84.44$\pm$2.73 & 84.61 & 0.908$\pm$0.022 & 0.915 \\
DBA L1(h=16) & 83.69$\pm$2.71 & 82.95 & 0.904$\pm$0.024 & 0.908 \\
DBA L2(h=1) & 79.91$\pm$2.81 & 80.79 & 0.885$\pm$0.026 & 0.903 \\
DBA L2(h=2) & \underline{85.20$\pm$2.90} & 84.10 & 0.900$\pm$0.029 & 0.899 \\
DBA L2(h=4) & 84.15$\pm$3.17 & 83.72 & 0.909$\pm$0.025 & 0.904 \\
DBA L2(h=8) & 84.28$\pm$2.66 & 84.99 & 0.906$\pm$0.023 & 0.917 \\
DBA L2(h=16) & 83.29$\pm$3.19 & 83.72 & \underline{0.911$\pm$0.024} & 0.923 \\
\bottomrule
\end{tabular}
\end{table}

\begin{table}[ht]
\centering
\caption{Validation and test performance across different attention mechanisms for pretrained ViT backbone}
\label{tab:vit_pre}
\begin{tabular}{lcccc}
\toprule
Experiment & Val Acc (\%) & Test Acc (\%) & Val AUC & Test AUC \\
\midrule
No attention(mean) & 76.74$\pm$2.46 & 74.65 & 0.838$\pm$0.023 & 0.819 \\
No attention(max) & 77.24$\pm$3.34 & 75.42 & 0.847$\pm$0.029 & 0.835 \\
VEMA(h=1) & 77.21$\pm$2.53 & 77.72 & 0.844$\pm$0.028 & 0.856 \\
VEMA(h=2) & 80.14$\pm$3.42 & 78.49 & 0.875$\pm$0.021 & 0.853 \\
VEMA(h=4) & 81.46$\pm$2.42 & 80.03 & 0.898$\pm$0.018 & 0.877 \\
VEMA(h=8) & 81.71$\pm$2.43 & 80.92 & 0.889$\pm$0.012 & 0.890 \\
VEMA(h=16) & 81.14$\pm$2.47 & 80.92 & 0.896$\pm$0.018 & 0.891 \\
DBA L1(h=1) & 76.94$\pm$1.99 & 75.80 & 0.837$\pm$0.022 & 0.822 \\
DBA L1(h=2) & 81.93$\pm$2.22 & 79.13 & 0.886$\pm$0.017 & 0.864 \\
DBA L1(h=4) & 81.81$\pm$2.65 & 80.79 & 0.896$\pm$0.015 & 0.888 \\
DBA L1(h=8) & 80.89$\pm$2.92 & 80.41 & 0.891$\pm$0.019 & 0.886 \\
DBA L1(h=16) & 81.49$\pm$2.51 & \underline{81.05} & \underline{0.900$\pm$0.020} & \underline{0.901} \\
DBA L2(h=1) & 76.74$\pm$2.68 & 77.46 & 0.841$\pm$0.028 & 0.846 \\
DBA L2(h=2) & 81.72$\pm$2.79 & 80.15 & 0.893$\pm$0.018 & 0.889 \\
DBA L2(h=4) & 81.82$\pm$2.54 & 78.87 & 0.893$\pm$0.016 & 0.866 \\
DBA L2(h=8) & \underline{81.90$\pm$2.31} & 76.95 & 0.891$\pm$0.019 & 0.863 \\
DBA L2(h=16) & 81.25$\pm$2.88 & 78.87 & 0.894$\pm$0.018 & 0.884 \\
\bottomrule
\end{tabular}
\end{table}

\begin{table}[ht]
\centering
\caption{Comparison of validation and test performance across different attention mechanisms for SimCLR using ViT pretrained backbone}
\label{tab:vit_simclr}
\begin{tabular}{lcccc}
\toprule
Experiment & Val Acc (\%) & Test Acc (\%) & Val AUC & Test AUC \\
\midrule
No attention(mean) & 78.49$\pm$3.47 & 80.92 & 0.861$\pm$0.026 & 0.882 \\
No attention(max) & 76.32$\pm$2.96 & 81.17 & 0.856$\pm$0.028 & 0.879 \\
VEMA(h=1) & 77.94$\pm$3.46 & 75.83 & 0.865$\pm$0.030 & 0.852 \\
VEMA(h=2) & 81.64$\pm$2.66 & 80.03 & 0.896$\pm$0.031 & 0.899 \\
VEMA(h=4) & \underline{83.00$\pm$1.96} & 80.41 & 0.905$\pm$0.016 & 0.886 \\
VEMA(h=8) & 81.75$\pm$2.71 & 82.57 & \underline{0.909$\pm$0.015} & 0.910 \\
VEMA(h=16) & 81.66$\pm$2.87 & 82.44 & 0.904$\pm$0.025 & \underline{0.914} \\
DBA L1(h=1) & 76.82$\pm$2.75 & 78.12 & 0.857$\pm$0.025 & 0.858 \\
DBA L1(h=2) & 82.13$\pm$2.41 & 82.06 & \underline{0.909$\pm$0.020} & 0.896 \\
DBA L1(h=4) & 82.16$\pm$2.96 & 78.88 & 0.904$\pm$0.022 & 0.874 \\
DBA L1(h=8) & 82.27$\pm$2.74 & 80.79 & 0.904$\pm$0.017 & 0.883 \\
DBA L1(h=16) & 81.80$\pm$3.18 & 79.52 & 0.904$\pm$0.022 & 0.895 \\
DBA L2(h=1) & 77.10$\pm$3.62 & 79.39 & 0.853$\pm$0.032 & 0.844 \\
DBA L2(h=2) & 81.78$\pm$2.00 & 79.01 & 0.892$\pm$0.028 & 0.854 \\
DBA L2(h=4) & 81.90$\pm$2.34 & \underline{83.72} & 0.903$\pm$0.018 & 0.896 \\
DBA L2(h=8) & 81.03$\pm$2.49 & 80.92 & 0.899$\pm$0.025 & 0.877 \\
DBA L2(h=16) & 82.31$\pm$2.28 & 82.57 & 0.907$\pm$0.020 & 0.910 \\
\hline
\end{tabular}
\end{table}

\begin{table}[ht]
\centering
\caption{Comparison of validation and test performance across different attention mechanisms for pretrained EfficientNet}
\label{tab:efficientnet_pre}
\begin{tabular}{lcccc}
\toprule
Experiment & Val Acc (\%) & Test Acc (\%) & Val AUC & Test AUC \\
\midrule
No attention(mean) & 76.32$\pm$3.35 & 76.19 & 0.845$\pm$0.031 & 0.845 \\
No attention(max) & 75.63$\pm$2.57 & 76.59 & 0.842$\pm$0.021 & 0.855 \\
VEMA(h=1) & 77.07$\pm$2.96 & 76.19 & 0.845$\pm$0.030 & 0.844 \\
VEMA(h=2) & 81.19$\pm$1.74 & 78.70 & 0.886$\pm$0.024 & 0.891 \\
VEMA(h=4) & 81.23$\pm$3.19 & \underline{81.88} & 0.882$\pm$0.029 & \underline{0.906} \\
VEMA(h=8) & 81.79$\pm$3.32 & 78.70 & 0.893$\pm$0.023 & 0.864 \\
VEMA(h=16) & 81.57$\pm$3.07 & 80.42 & 0.894$\pm$0.022 & 0.883 \\
DBA L1(h=1) & 76.48$\pm$3.66 & 76.06 & 0.838$\pm$0.031 & 0.844 \\
DBA L1(h=2) & 81.75$\pm$2.91 & 79.76 & 0.890$\pm$0.029 & 0.885 \\
DBA L1(h=4) & \underline{81.90$\pm$2.48} & 80.29 & 0.888$\pm$0.024 & 0.867 \\
DBA L1(h=8) & 82.01$\pm$2.72 & 78.04 & 0.891$\pm$0.018 & 0.859 \\
DBA L1(h=16) & 81.30$\pm$1.89 & 79.89 & \underline{0.893$\pm$0.024} & 0.877 \\
DBA L2(h=1) & 76.57$\pm$2.35 & 79.50 & 0.841$\pm$0.023 & 0.870 \\
DBA L2(h=2) & 80.22$\pm$3.54 & 77.78 & 0.871$\pm$0.025 & 0.868 \\
DBA L2(h=4) & 81.40$\pm$3.62 & 79.76 & 0.882$\pm$0.034 & 0.877 \\
DBA L2(h=8) & 81.41$\pm$2.43 & 79.23 & 0.886$\pm$0.026 & 0.863 \\
DBA L2(h=16) & 81.15$\pm$2.97 & 80.82 & 0.889$\pm$0.031 & 0.858 \\
\bottomrule
\end{tabular}
\end{table}

\begin{table}[ht]
\centering
\caption{Comparison of validation and test performance across different attention mechanisms using SimCLR and pretrained Efficientnet as backbone}
\label{tab:efficientnet_simclr}
\begin{tabular}{lcccc}
\toprule
Experiment & Val Acc (\%) & Test Acc (\%) & Val AUC & Test AUC \\
\midrule
No attention(mean) & 79.01$\pm$3.99 & 81.42 & 0.874$\pm$0.029 & 0.900 \\
No attention(max) & 80.11$\pm$3.21 & 80.66 & 0.876$\pm$0.028 & 0.886 \\
VEMA(h=1) & 80.15$\pm$3.57 & 79.90 & 0.880$\pm$0.030 & 0.899 \\
VEMA(h=2) & 83.53$\pm$2.81 & 82.57 & 0.902$\pm$0.034 & 0.901 \\
VEMA(h=4) & 83.99$\pm$3.18 & 81.68 & 0.899$\pm$0.027 & 0.890 \\
VEMA(h=8) & 83.30$\pm$2.99 & 83.33 & 0.901$\pm$0.024 & 0.894 \\
VEMA(h=16) & 82.75$\pm$3.46 & 79.64 & 0.907$\pm$0.026 & 0.883 \\
DBA L1(h=1) & 80.49$\pm$2.67 & 80.66 & 0.882$\pm$0.027 & 0.888 \\
DBA L1(h=2) & \underline{84.82$\pm$2.22} & 83.46 & 0.903$\pm$0.024 & \underline{0.926} \\
DBA L1(h=4) & 83.63$\pm$2.97 & 82.44 & 0.896$\pm$0.029 & 0.885 \\
DBA L1(h=8) & 84.01$\pm$2.97 & 80.92 & 0.901$\pm$0.028 & 0.900 \\
DBA L1(h=16) & 83.12$\pm$2.74 & 83.84 & 0.901$\pm$0.027 & 0.925 \\
DBA L2(h=1) & 79.49$\pm$3.06 & 80.28 & 0.876$\pm$0.030 & 0.878 \\
DBA L2(h=2) & 83.46$\pm$2.59 & 83.72 & 0.897$\pm$0.028 & 0.906 \\
DBA L2(h=4) & 83.71$\pm$3.20 & \underline{83.97} & 0.900$\pm$0.022 & 0.910 \\
DBA L2(h=8) & \underline{84.82$\pm$3.01} & 81.55 & \underline{0.911$\pm$0.022} & 0.894 \\
DBA L2(h=16) & 83.58$\pm$2.17 & 82.06 & 0.907$\pm$0.021 & 0.892 \\
\bottomrule
\end{tabular}
\end{table}

\begin{table}[ht]
\centering
\caption{Comparison of validation and test performance across different attention mechanisms for pretrained ConvNeXt}
\label{tab:ConvNeXt_pre}
\begin{tabular}{lcccc}
\toprule
Experiment & Val Acc (\%) & Test Acc (\%) & Val AUC & Test AUC \\
\midrule
No attention(mean) & 79.99$\pm$2.55 & 81.68 & 0.879$\pm$0.026 & 0.895 \\
No attention(max) & 79.64$\pm$2.35 & 80.07 & 0.878$\pm$0.018 & 0.887 \\
VEMA(h=1) & 80.43$\pm$3.06 & 81.81 & 0.885$\pm$0.019 & 0.888 \\
VEMA(h=2) & 82.30$\pm$2.00 & 82.05 & 0.898$\pm$0.024 & 0.901 \\
VEMA(h=4) & 82.72$\pm$2.10 & 83.29 & 0.904$\pm$0.016 & 0.903 \\
VEMA(h=8) & \underline{83.44$\pm$2.55} & 82.43 & 0.902$\pm$0.024 & 0.890 \\
VEMA(h=16) & 82.81$\pm$2.17 & 83.29 & 0.906$\pm$0.016 & 0.894 \\
DBA L1(h=1) & 80.06$\pm$3.26 & 80.32 & 0.876$\pm$0.019 & 0.886 \\
DBA L1(h=2) & 83.22$\pm$2.23 & 82.80 & 0.903$\pm$0.017 & 0.887 \\
DBA L1(h=4) & 83.25$\pm$2.19 & 83.29 & 0.909$\pm$0.019 & 0.901 \\
DBA L1(h=8) & 82.26$\pm$1.90 & 81.06 & 0.891$\pm$0.021 & 0.896 \\
DBA L1(h=16) & 83.24$\pm$2.61 & 82.05 & \underline{0.910$\pm$0.021} & 0.901 \\
DBA L2(h=1) & 80.86$\pm$2.97 & 81.68 & 0.886$\pm$0.019 & 0.894 \\
DBA L2(h=2) & 82.53$\pm$2.56 & \underline{83.66} & 0.895$\pm$0.029 & \underline{0.913} \\
DBA L2(h=4) & 83.18$\pm$1.74 & 81.68 & 0.900$\pm$0.020 & 0.889 \\
DBA L2(h=8) & 83.04$\pm$1.83 & 83.04 & 0.895$\pm$0.020 & 0.894 \\
DBA L2(h=16) & 82.88$\pm$1.97 & 81.31 & 0.905$\pm$0.015 & 0.904 \\
\bottomrule
\end{tabular}
\end{table}

\begin{table}[ht]
\centering
\caption{Comparison of validation and test performance across different attention mechanisms for SimCLR using pretrained ConvNeXt backbone}
\label{tab:ConvNeXt_simclr}
\begin{tabular}{lcccc}
\toprule
Experiment & Val Acc (\%) & Test Acc (\%) & Val AUC & Test AUC \\
\midrule
No attention(mean) & 81.63$\pm$3.84 & 82.95 & 0.900$\pm$0.027 & 0.910 \\
No attention(max) & 81.69$\pm$2.51 & 81.55 & 0.899$\pm$0.021 & 0.911 \\
VEMA(h=1) & 81.20$\pm$2.99 & 81.93 & 0.897$\pm$0.020 & 0.897 \\
VEMA(h=2) & \underline{85.86$\pm$3.05} & 82.32 & \underline{0.915$\pm$0.028} & 0.882 \\
VEMA(h=4) & 84.84$\pm$2.80 & 83.33 & 0.911$\pm$0.028 & 0.908 \\
VEMA(h=8) & 84.13$\pm$3.09 & 84.10 & 0.907$\pm$0.026 & 0.913 \\
VEMA(h=16) & 83.63$\pm$3.67 & 83.46 & 0.907$\pm$0.025 & 0.901 \\
DBA L1(h=1) & 81.68$\pm$3.14 & 81.55 & 0.896$\pm$0.016 & 0.904 \\
DBA L1(h=2) & 84.54$\pm$3.19 & \underline{86.26} & 0.914$\pm$0.021 & \underline{0.928} \\
DBA L1(h=4) & 84.98$\pm$2.73 & 85.24 & 0.910$\pm$0.027 & 0.907 \\
DBA L1(h=8) & 84.75$\pm$2.77 & 83.46 & 0.913$\pm$0.024 & 0.920 \\
DBA L1(h=16) & 84.39$\pm$2.78 & 83.08 & 0.911$\pm$0.021 & 0.903 \\
DBA L2(h=1) & 82.07$\pm$3.36 & 83.08 & 0.901$\pm$0.020 & 0.911 \\
DBA L2(h=2) & 85.66$\pm$3.29 & 84.48 & 0.914$\pm$0.030 & 0.922 \\
DBA L2(h=4) & 84.99$\pm$3.63& 83.33 & 0.913$\pm$0.032 & 0.920 \\
DBA L2(h=8) & 83.99$\pm$3.14 & 84.10 & 0.912$\pm$0.025 & 0.904 \\
DBA L2(h=16) & 84.11$\pm$2.46 & 84.10 & 0.909$\pm$0.022 & 0.908 \\
\bottomrule
\end{tabular}
\end{table}

\begin{figure}[h]
    \centering
    \begin{tabular}{c c}
        True Positive  & False Negative  \\
        \includegraphics[width=0.45\textwidth]{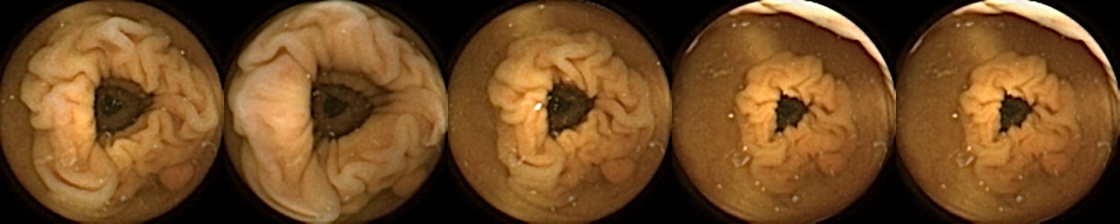} &
        \includegraphics[width=0.45\textwidth]{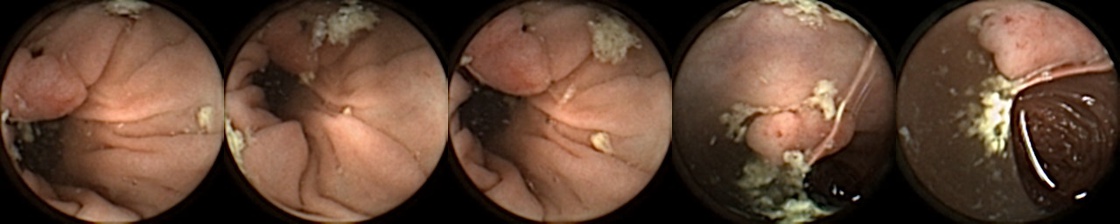}
        \\
        \includegraphics[width=0.45\textwidth]{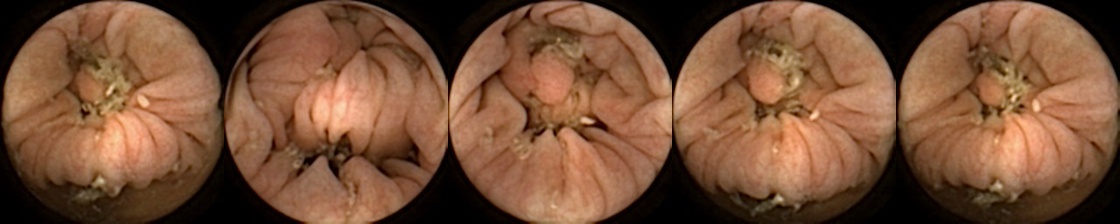}  &
        \includegraphics[width=0.45\textwidth]{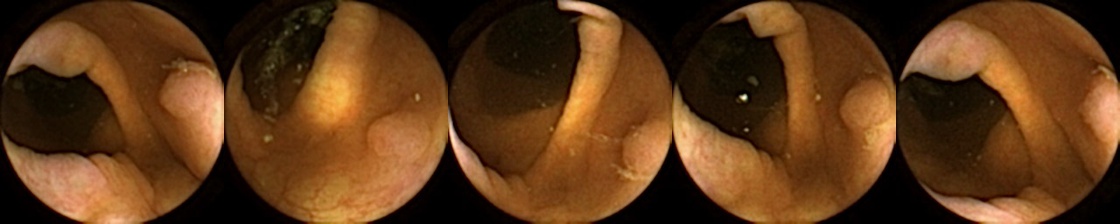}
        \\
        \includegraphics[width=0.45\textwidth]{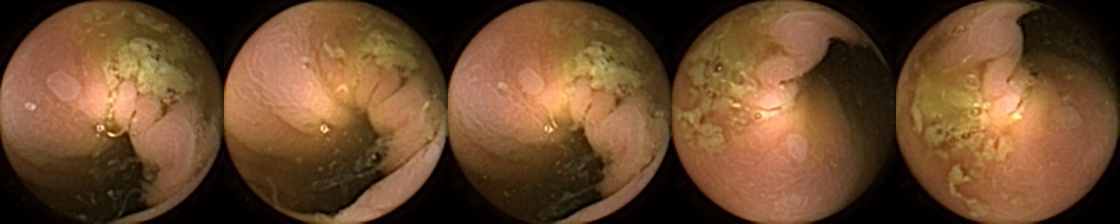}  &
        \includegraphics[width=0.45\textwidth]{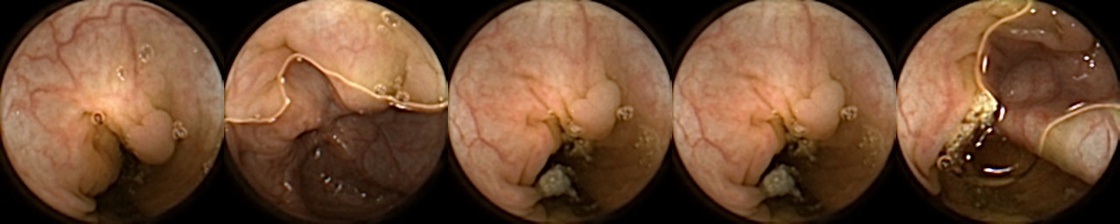}
        \\
        \includegraphics[width=0.45\textwidth]{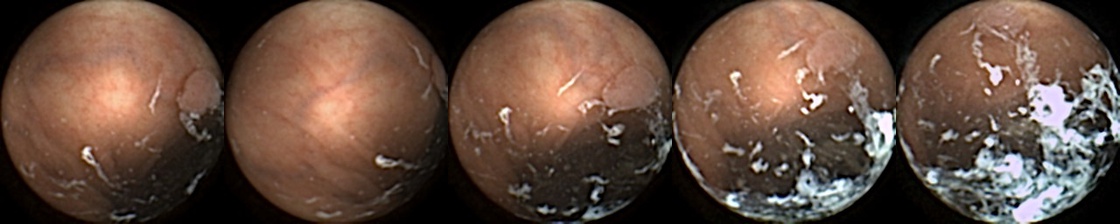}  &
        \includegraphics[width=0.45\textwidth]{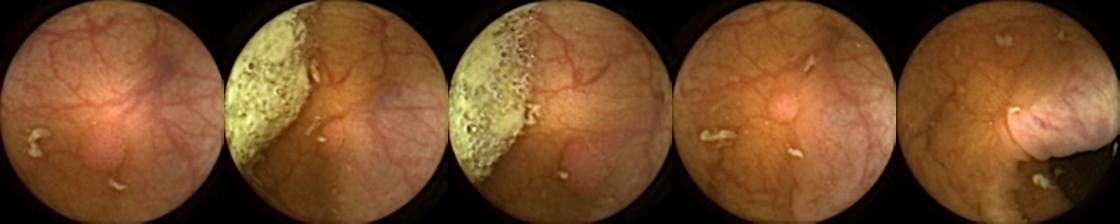}
        \\
        \includegraphics[width=0.45\textwidth]{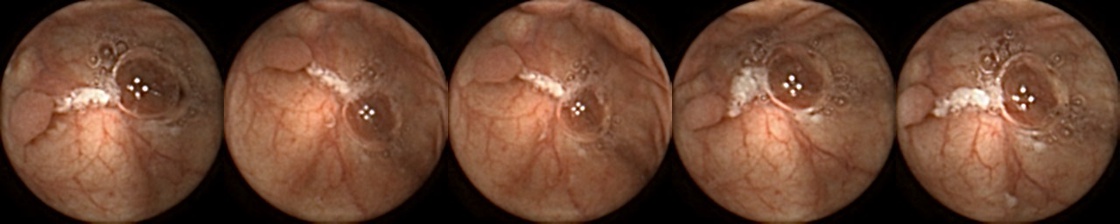}  &
        \includegraphics[width=0.45\textwidth]{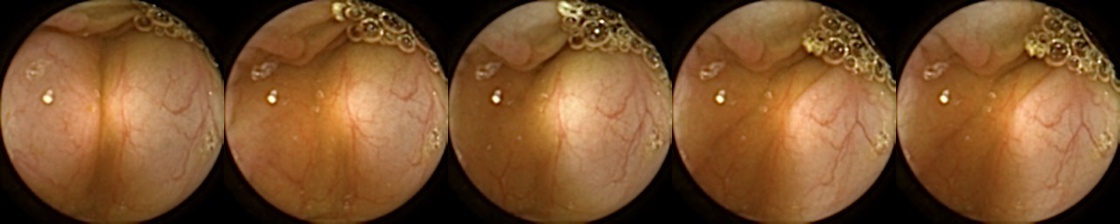}
        \\
        False Positive & True Negative \\
         \includegraphics[width=0.45\textwidth]{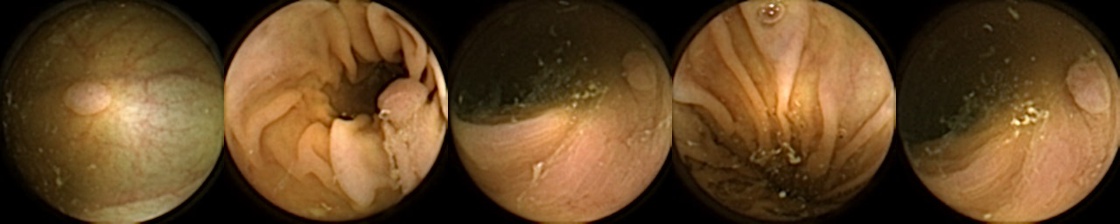} &
        \includegraphics[width=0.45\textwidth]{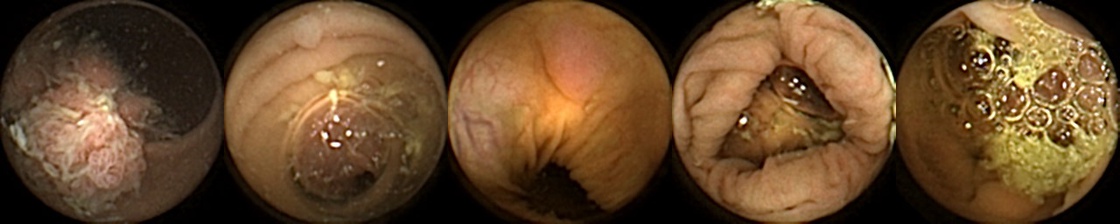}
        \\
        \includegraphics[width=0.45\textwidth]{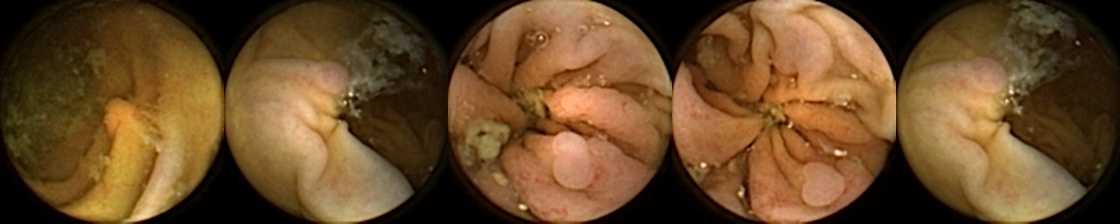}  &
        \includegraphics[width=0.45\textwidth]{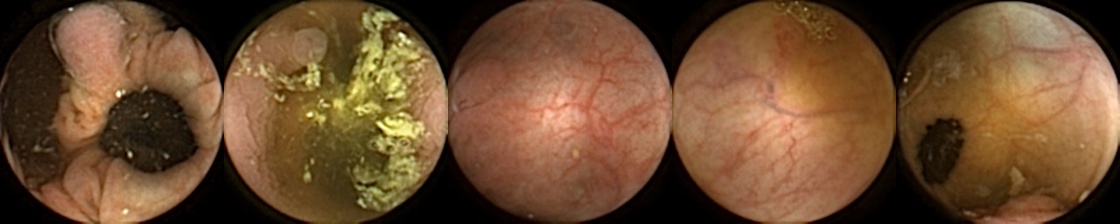}
        \\
        \includegraphics[width=0.45\textwidth]{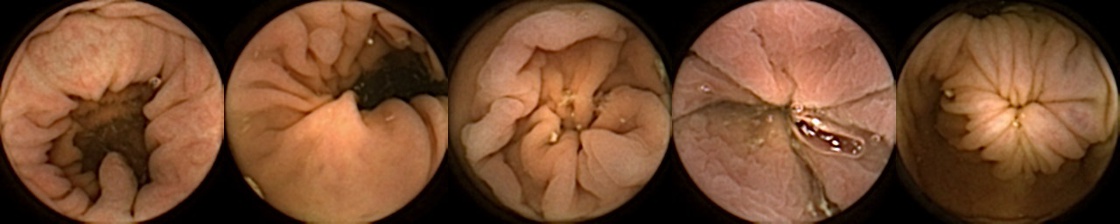}  &
        \includegraphics[width=0.45\textwidth]{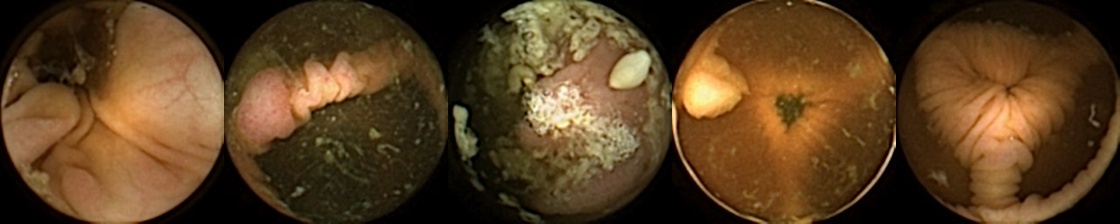}
        \\
        \includegraphics[width=0.45\textwidth]{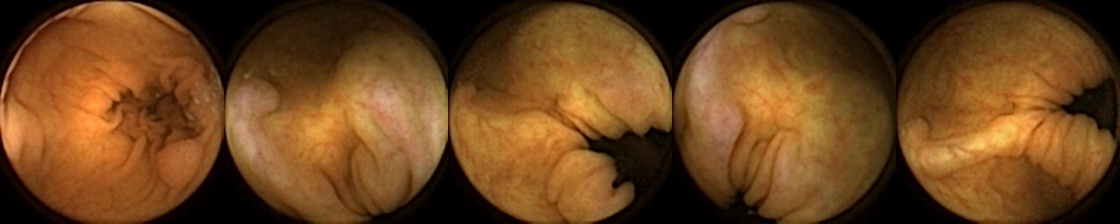}  &
        \includegraphics[width=0.45\textwidth]{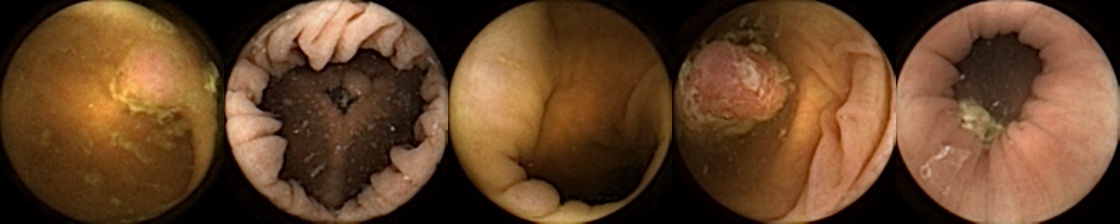}
        \\
        \includegraphics[width=0.45\textwidth]{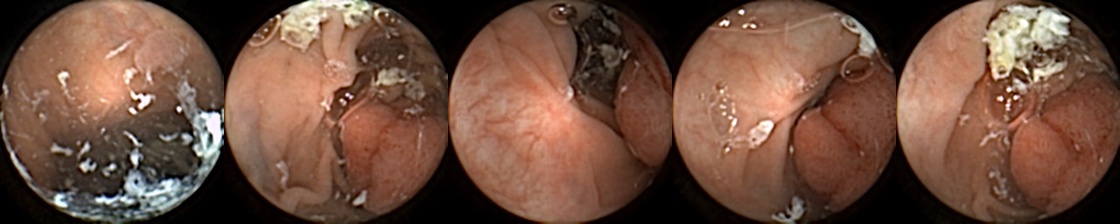}  &
        \includegraphics[width=0.45\textwidth]{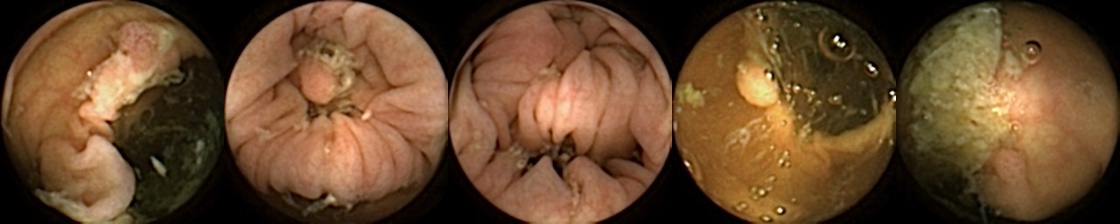}
        \\
        
    \end{tabular}
    \caption{True Positives (Pred = true, Label = true), False Negatives (Pred = false, Label = true), False Positives (Pred = true, Label = false), True Negatives (Pred = false, Label = false) for the DBA L2(h=2) model using the pretrained ConvNeXt. In each row, the leftmost image is the query and the 4 images to the right of each query are the target images.}
    \label{fig:convnext_pre}
   
\end{figure}

Figure~\ref{fig:convnext_pre} shows classification output examples for the MIV model using the pretrained ConvNext. Note that the model has a test accuracy of 83.66\% for DBA L2 with 2 heads.  In each row, the leftmost image is the query and the four images to the right of each query are the target images. The True Positive (left-top) and True Negative (bottom-right) examples outline the cases when the model is able to successfully distinguishes images containing unique polyps from those with dissimilar polyps. The False Negative (top-right) and False Positive (bottom-left) examples show the cases where the model fails to correctly distinguish images of unique polyps and images of dissimilar polyps. Further, we can note from the differences in the examples of True Positives and False Negatives that when the query image and the target set instances differ from each other in some ways, then the model can misclassify images. These differences could be attributed to different views due to the dynamics of the camera inside the colon, different views from two camera heads of the capsule, or presence of artifacts such as bubbles, debris, and small bowel secretions. 



On the other hand, when we look at the examples of False Positives (Pred = true, Label = false) in Figure~\ref{fig:convnext_pre}, we can note that if images in the target match the query in ways such as texture, color, illumination conditions, presence of artifacts, then this may lead to misidentification of the query as belonging to the targets. For True Negative (Pred = false, Label = false) examples, we can see that presence of significant differences between the query and the targets leads to correct identification of the image sets as different.

\begin{figure}[h]
    \centering
    \begin{tabular}{c c}
        True Positive  & False Negative  \\
        \includegraphics[width=0.45\textwidth]{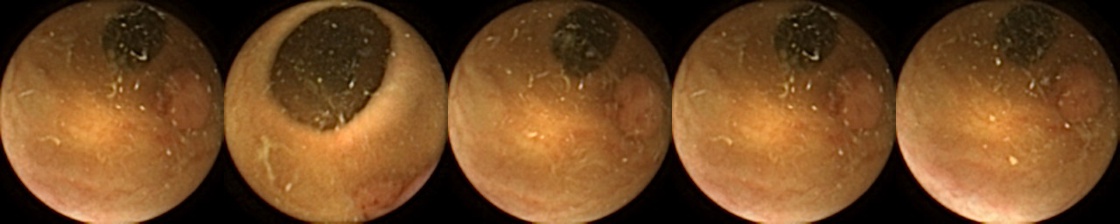} &
        \includegraphics[width=0.45\textwidth]{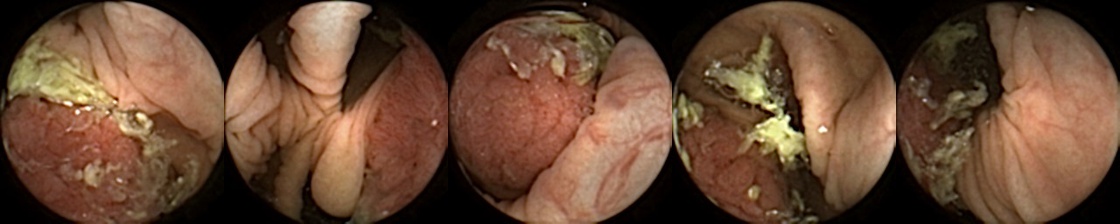}
        \\
        \includegraphics[width=0.45\textwidth]{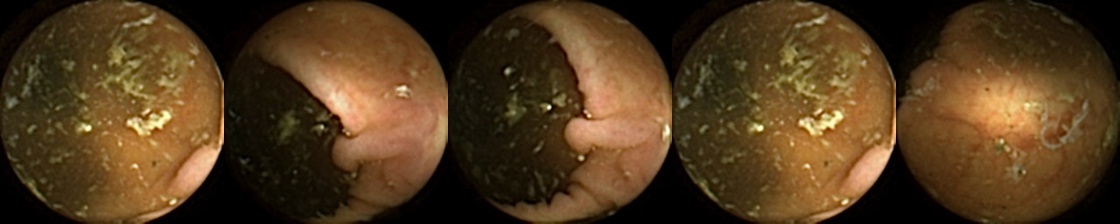}  &
        \includegraphics[width=0.45\textwidth]{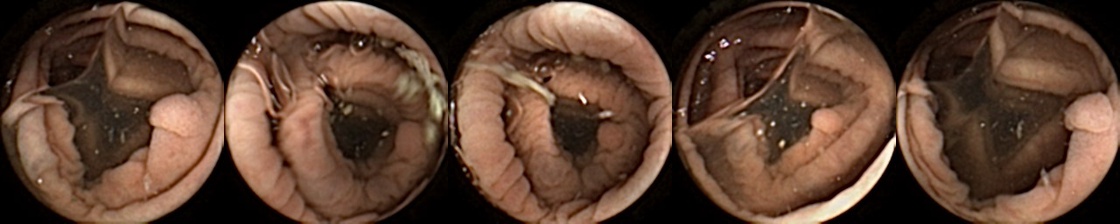}
        \\
        \includegraphics[width=0.45\textwidth]{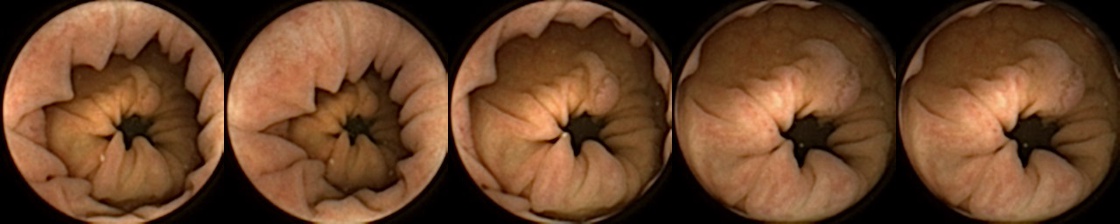}  &
        \includegraphics[width=0.45\textwidth]{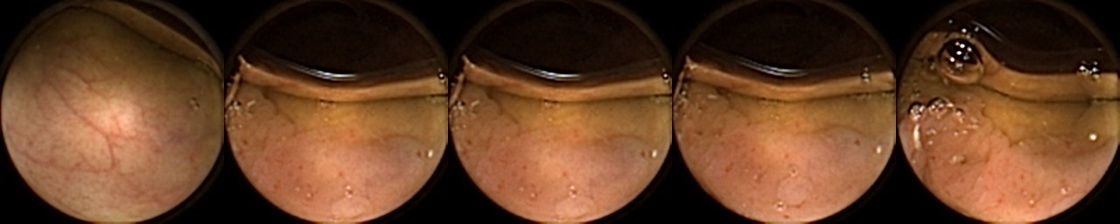}
        \\
        \includegraphics[width=0.45\textwidth]{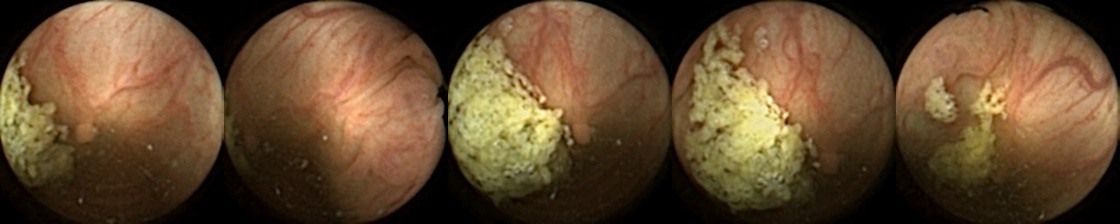}  &
        \includegraphics[width=0.45\textwidth]{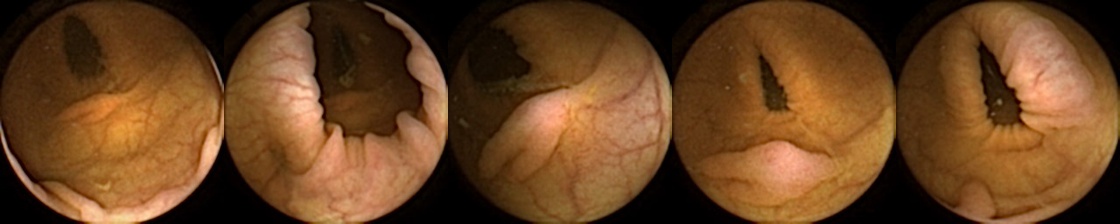}
        \\
        \includegraphics[width=0.45\textwidth]{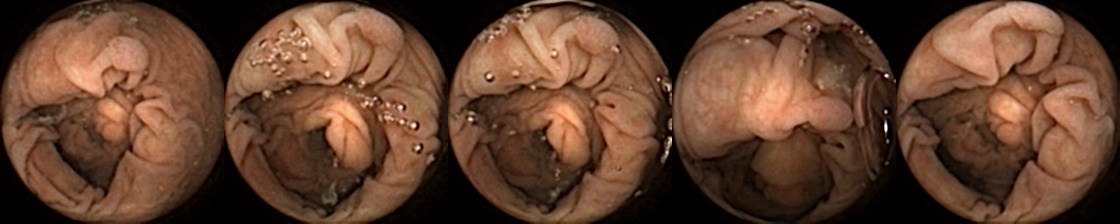}  &
        \includegraphics[width=0.45\textwidth]{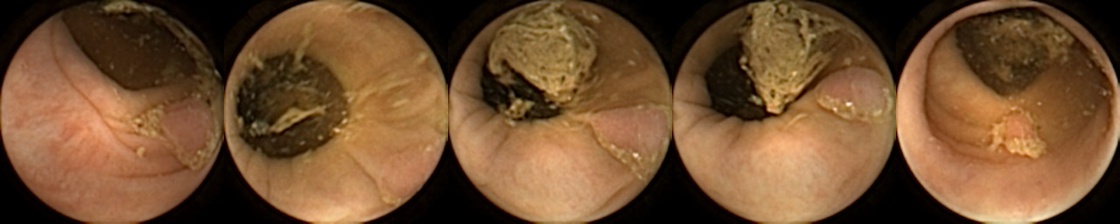}
        \\
        False Positive & True Negative \\
         \includegraphics[width=0.45\textwidth]{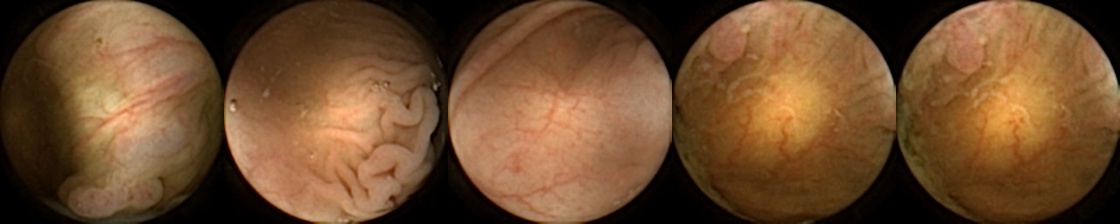} &
        \includegraphics[width=0.45\textwidth]{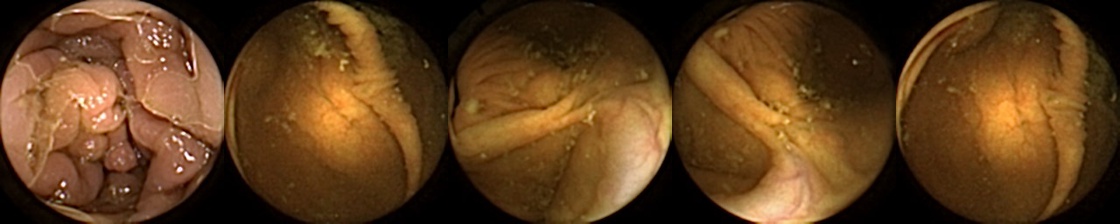}
        \\
        \includegraphics[width=0.45\textwidth]{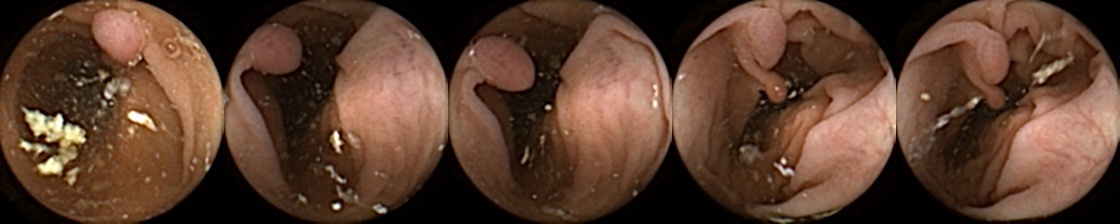}  &
        \includegraphics[width=0.45\textwidth]{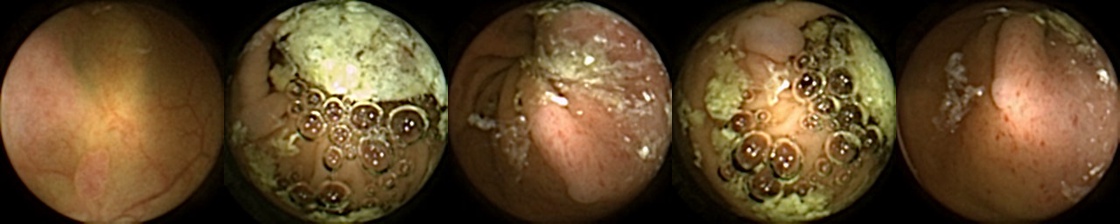}
        \\
        \includegraphics[width=0.45\textwidth]{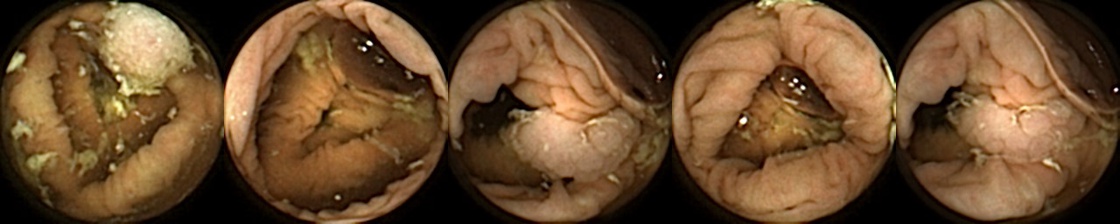}  &
        \includegraphics[width=0.45\textwidth]{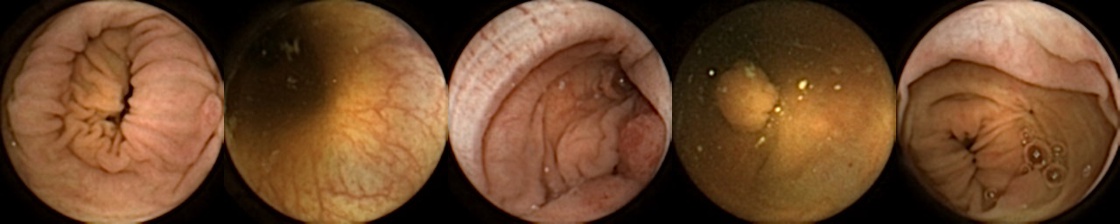}
        \\
        \includegraphics[width=0.45\textwidth]{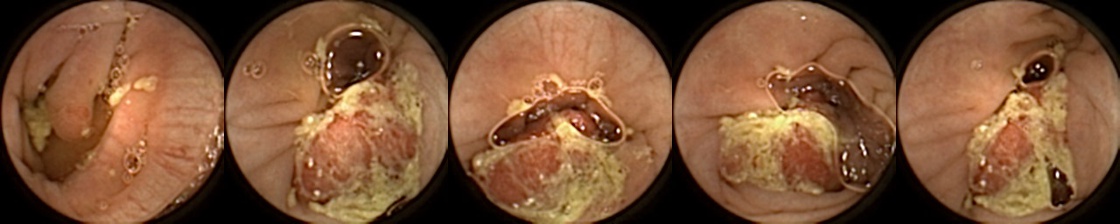}  &
        \includegraphics[width=0.45\textwidth]{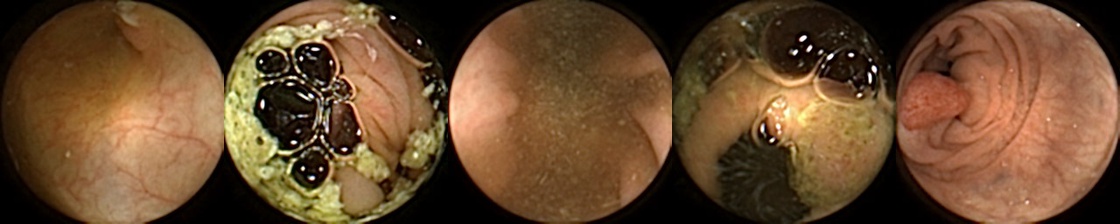}
        \\
        \includegraphics[width=0.45\textwidth]{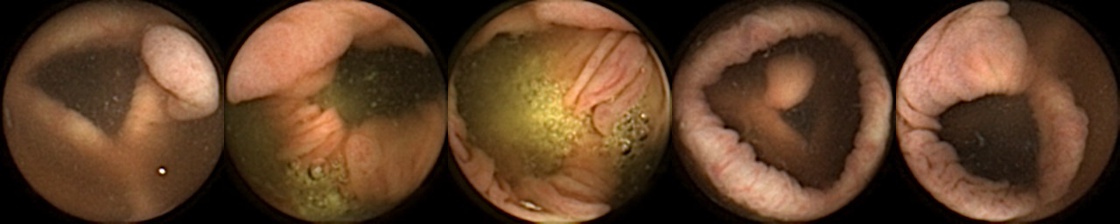}  &
        \includegraphics[width=0.45\textwidth]{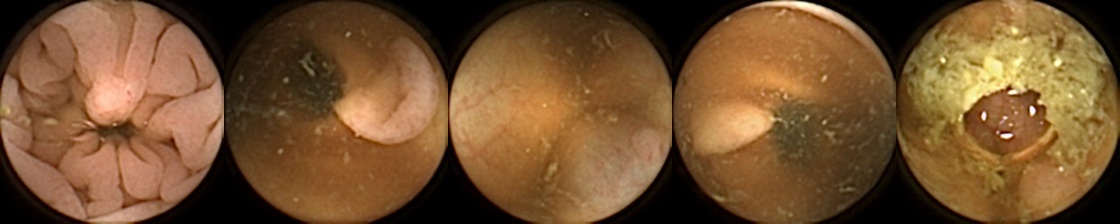}
        \\
        
    \end{tabular}
    \caption{True Positives (Pred = true, Label = true), False Negatives (Pred = false, Label = true), False Positives (Pred = true, Label = false),True Negatives (Pred = false, Label = false) for the DBA L1(h=2) model applying SimCLR using the ConvNeXt backbone. In each row, the leftmost image is the query and the 4 images to the right of each query are the target images.}
\label{fig:convnext_simclr}
   
\end{figure}

Similarly, we consider examples for the MIV model based on SimCLR pretraining using the pretrained ConvNext backbone in Figure~\ref{fig:convnext_simclr}. The model has a test accuracy of 86.26\% for DBA L1 with 2 heads. Here we note again a similar trend by looking at the True Positive and False Negative examples as observed before, where a significant difference in the query image when compared with the target images can lead to a failure to identify the same polyp images in the five examples. A look at the False Positive examples affirms that the task remains challenging, as evidenced by the model's occasional misclassifications in ambiguous cases.

\begin{figure}[h]
    \centering
        \includegraphics[width=0.85\textwidth]{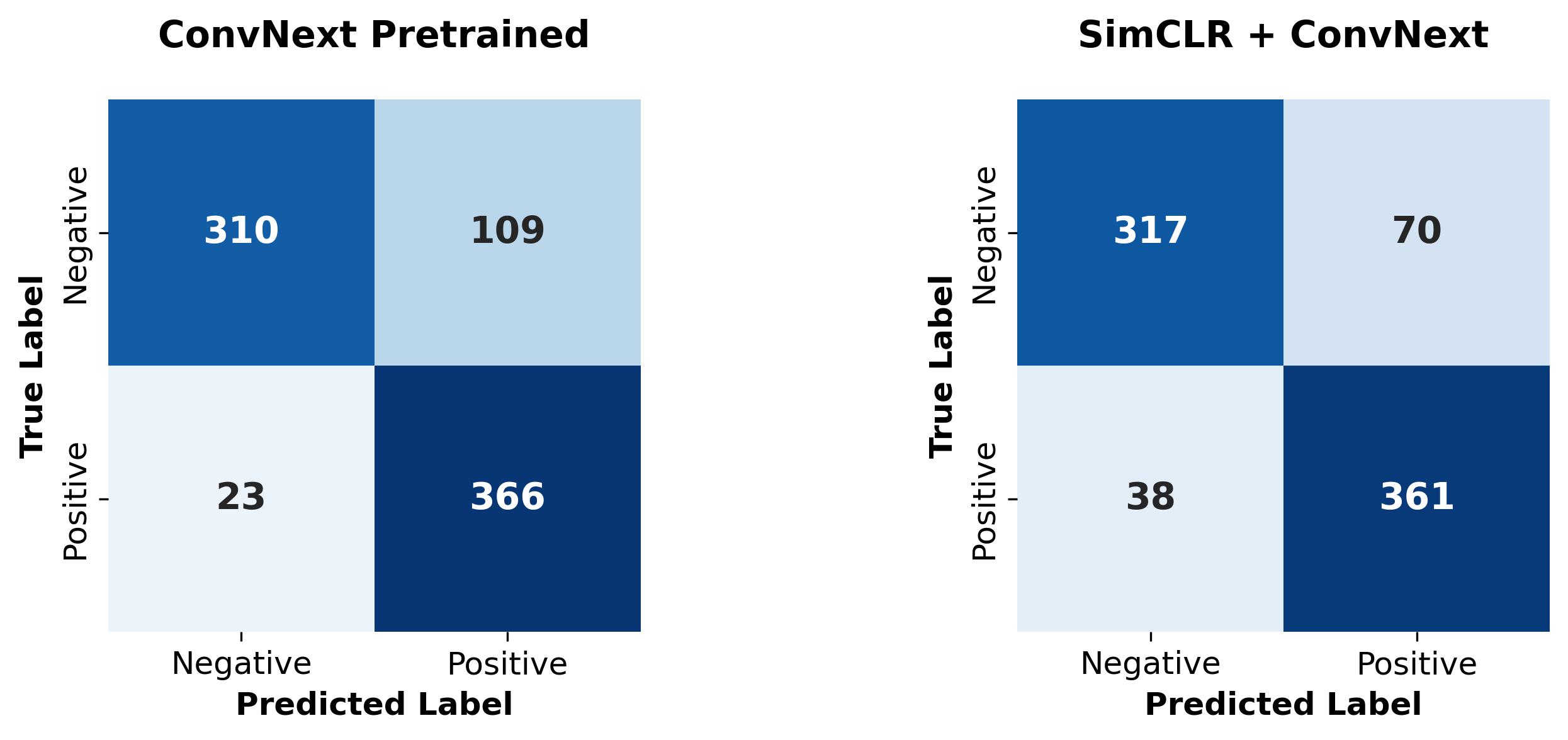} 
\label{fig:confusion_mats}
\caption{Confusion matrices from the best MIV model associated with (left) the pretrained ConvNext, and (right) SimCLR pretraining using the ConvNext backbone } 
\end{figure}

We can additionally consider the confusion matrices from the best MIV model associated with (a) the pretrained ConvNext and (b) SimCLR pretraining using the ConvNext backbone in Figure~\ref{fig:confusion_mats}. We note that using SimCLR achieves a better ratio of False Negatives (70) to False Positives (38) as compared to pretrained ConvNext, which has 109 False Negatives and 23 False Positives.


\subsection{Discussion}


Based on the results presented in the study, the proposed Multi-Instance Learning (MIL) framework demonstrates significant potential in addressing the challenge of identifying unique polyps in \ac{CCE} images. The integration of attention mechanisms, particularly Distance-Based Attention (DBA) and Variance-Excited Multi-head Attention (VEMA), has shown to enhance the model's ability to extract useful representations, leading to improved classification performance. Notably, the DBA L1 mechanism with SimCLR pretraining achieved the highest test accuracy of 86.26\% and a test AUC of 0.928, underscoring the efficacy of combining self-supervised learning with attention-based MIL frameworks. However, the analysis of misclassifications of positive examples (False Negatives) reveals that the task remains challenging, particularly in cases where the query and target images exhibit significant differences. On the other hand, in case of False Positives, we note that the presence of similar structures (texture, color, or artifacts) between the query image and target bag can lead to misidentification of different polyp images as belonging to the same polyp.

The misclassifications (False Negatives and False Positives) in MIV models can be attributed to first, the fact that the dataset is limited to 1912 unique polyps from 754 patients and both, studies and data, within this specific subdomain are lacking. The MIV models can struggle to classify images that differ significantly from those seen during the training phase. Second, the PillCam capsule with its two-headed camera can capture different views of a polyp sometimes from opposite directions hence making the task of identifying unique polyps quite challenging. 


To alleviate the challenges of a lack of data, use of synthetic data generation as discussed in~\cite{gatoula2025clinical} for CCE images can be seen as the next step to address the challenges of limited data with labels. However, according to the authors, it is important to note that synthetic images can have unrealistic appearance of anatomical structures~\cite{gatoula2025clinical}.

We believe that our proposed methodology can be extended to other diagnoses (e.g., ulcers, bleeding, diverticula, and inflammatory bowel disease)~\cite{hosoe2020colon, thorndal2025colon} within CCE and even domains beyond CCE,  where large datasets with limited or weakly labeled instances are common. By replacing the reliance on extensive manual annotations to annotations at bag level, this work paves the way for more efficient, accurate, and scalable diagnostic tools. Ultimately, these advancements can contribute to reducing clinician workload, improving diagnostic accuracy, thereby enhancing patient outcomes and advancing the broader field of medical imaging and AI.

\section{Conclusions}

 In this paper, we formulate the problem of identifying unique polyps in CCE images as a multi-instance learning problem where a query polyp image is compared with a target bag of images.
 Baseline models without attention mechanisms, utilizing mean or max pooling strategies, achieve moderate performance, with validation accuracies ranging from 76.3\% to 81.7\% and test accuracies between 74.2\% and 82.9\%. The inclusion of attention mechanisms, such as VEMA and Distance-Based Attention (DBA), lead to substantial improvements. VEMA shows strong gains, particularly with multi-head configurations, while DBA mechanisms, leveraging L1 and L2 distance metrics, often outperform VEMA and achieve competitive results across architectures. Notably, DBA L1 with 2 heads achieves the highest test accuracy of 86.26\% and a test AUC of 0.928 using the ConvNeXt backbone with SimCLR pretraining. SimCLR pretraining further enhances performance, with models pretrained using SimCLR consistently achieving higher validation and test accuracies compared to standard pretrained counterparts. Despite these advancements, challenges remain in distinguishing between images of unique polyps and dissimilar polyps, particularly in edge cases where query images and target instances differ, underscoring the complexity of the classification task.

\section*{Acknowledgment}
This research is part of AICE project (number 101057400) funded by the European Union, and it is part-funded by the United Kingdom government. Views and opinions expressed are however those of the author(s) only and do not necessarily reflect those of the European Union or the European Commission. Neither the European Union nor the European Commission can be held responsible for them.
\begin{figure}[H]
    \centering
    \includegraphics[height=1.1cm]{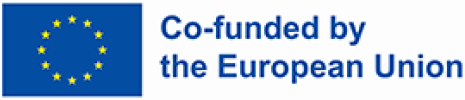}
    \includegraphics[height=1.1cm]{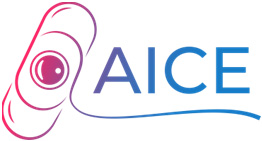}
    \label{fig:funding}
\end{figure}

\bibliography{mybibliography}
\end{document}